\pdfoutput=1
\documentclass[sigconf,nonacm]{acmart}

\usepackage{amsmath}
\usepackage{booktabs}
\usepackage{graphicx}
\usepackage{textcomp}
\usepackage{xcolor}
\usepackage{dblfloatfix}
\usepackage{placeins}
\graphicspath{{./}{figures/}}
\title{PPO-HRAP: Proximal Policy Optimization with a Hybrid Regime-Aware Policy for Risk-Controlled Trading}

\author{Duong Hien Chi Kien}
\email{24kien.dhc@vinuni.edu.vn}
\affiliation{%
  \institution{College of Engineering and Computer Science, VinUniversity}
  \city{Ha Noi}
  \country{Viet Nam}
}

\author{Huynh Thanh Trung}
\email{trung.ht@vinuni.edu.vn}
\affiliation{%
  \institution{College of Engineering and Computer Science, VinUniversity}
  \city{Ha Noi}
  \country{Viet Nam}
}

\begin{document}

\begin{abstract}
Reinforcement learning for trading often struggles to balance upside
participation with drawdown control. Profit-only policies can collapse toward
passive long exposure on upward-drifting assets, while aggressively
risk-penalized rewards can become too defensive during volatile periods. This
paper proposes PPO-HRAP, a hybrid regime-aware policy that combines Proximal
Policy Optimization with an interpretable regime prior. The agent observes both
market features and portfolio-state variables, receives a reward combining
portfolio log return, VIX-conditioned drawdown-increase penalty,
target-exposure deviation, and turnover cost, and executes a blended action
between the PPO actor output and a regime-derived target exposure. On the
held-out 2020--2022 SPY test window, PPO-HRAP achieves 27.62\% total return,
8.48\% annualized return, 0.6447 Sharpe ratio, 0.8588 Sortino ratio, and
0.4592 Calmar ratio, while reducing maximum drawdown from 34.10\% for Buy and
Hold to 18.47\%. Across five SPY seeds, PPO-HRAP remains stable with mean
total return $0.2725 \pm 0.0109$ and mean Sharpe ratio $0.6219 \pm 0.0565$.
Single-run cross-asset tests on QQQ and DIA further show that the proposed
method ranks first on total return and Sharpe ratio for all three reported
assets. These results suggest that blending learned actions with a
volatility-aware regime prior is a practical way to improve risk-adjusted
trading behavior, although the current policy still incurs high turnover and
cross-asset robustness beyond SPY remains limited to single-run evidence.
\end{abstract}


\maketitle
\raggedbottom

\section{Introduction}
Reinforcement learning (RL) has become a practical tool for portfolio management because it can optimize sequential allocation decisions directly from market interaction rather than from a fixed predictive model \cite{jiang2017portfolio,gu2025mts}. In trading, however, the reward function is often misaligned with the real objective. A policy trained only to maximize return can converge to behavior that is close to remaining fully long on an upward-drifting equity index. Conversely, a policy penalized too aggressively for risk can become excessively defensive and sacrifice most upside.

This tension is not merely an implementation detail. Daily allocation is a
closed-loop control problem: the action changes the portfolio state, the new
portfolio state changes future risk exposure, and the reward observed by the
agent mixes market movement with the consequences of prior trades. A useful
trading agent therefore needs both market information, such as trend and
volatility, and portfolio information, such as current weight, cash ratio, and
drawdown. If the state omits these portfolio variables, the agent receives an
incomplete description of the system it is controlling. If the reward ignores
drawdown or turnover, the agent can learn policies that look profitable in
average return but are unattractive as trading systems.

This problem becomes sharper when risk control depends on volatility signals such as the CBOE Volatility Index (VIX), a widely used indicator of market stress \cite{whaley2009vix}. VIX can indicate how aggressively a portfolio should be sized, but it does not by itself provide a strong directional prior. In a single-asset SPY plus cash setting, this creates an unstable design space: mild penalties fail to move the policy far enough away from passive exposure, whereas strong penalties may push it toward near-flat or short-biased behavior.

We address this gap with PPO-HRAP, short for PPO-Hybrid Regime-Aware Policy. The method uses a PPO backbone \cite{schulman2017ppo} but augments it with an explicit regime-aware action prior. The core design separates two questions: which direction the portfolio should lean, and how much risk should be taken under the current market regime. Trend signals define a desired directional exposure, while the VIX z-score controls drawdown sensitivity and exposure reduction under stress. Empirically, we evaluate the method with a main SPY benchmark, a five-seed SPY reliability study, and single-run transfers to QQQ and DIA.

The method is intentionally simple. Rather than asking a neural policy to
discover all regime structure only from delayed rewards, PPO-HRAP computes an
interpretable target exposure from trend and stress indicators and then blends
that target with the PPO action. This keeps the learned policy in the loop but
anchors execution to a prior that is easy to inspect. The design is therefore
closer to a hybrid control system than to a fully unconstrained black-box
trader.

The contribution is practical rather than purely theoretical. PPO-HRAP is designed as a hybrid control-plus-RL trading policy rather than as a purely learned reward innovation. Its main contributions are:
\begin{itemize}
    \item a hybrid reward formulation that combines portfolio log return, dynamic drawdown control, target-exposure regularization, and turnover control;
    \item an explicit regime-classification rule that separates directional guidance from volatility-aware risk sizing;
    \item an action-blending mechanism that treats the PPO policy as a correction around an interpretable prior instead of as an unconstrained controller; and
    \item a benchmarked evaluation pipeline spanning financial baselines, RL reward baselines, validation-based action-prior selection, SPY multi-seed reliability analysis, and cross-asset transfer to QQQ and DIA.
\end{itemize}

On the 2020--2022 test window, PPO-HRAP achieves the highest total return and
the highest Sharpe, Sortino, and Calmar ratios among all compared methods in
the main SPY run, while also transferring favorably to QQQ and DIA. The
multi-seed analysis adds nuance to this picture: averaged over seeds, SAC with
a profit-only reward attains better risk ratios at a lower total return. The
rest of the paper presents the related literature, the hybrid policy design,
the experimental protocol, and the resulting empirical trade-offs.

\section{Related Work}
Deep RL for trading and portfolio allocation has evolved from early direct reinforcement formulations into policy-based systems that operate on historical market features \cite{moody1998performance,moody2001learning,nevmyvaka2006rltrading,deng2017deepdirect,almahdi2017adaptive,liang2018adversarial,ye2020reinforcement,jiang2017portfolio,liu2020finrl,yang2020ensemble,liu2021finrl,liu2022finrlmeta,lucarelli2020crypto,millea2021criticalsurvey,jiang2024drlportfolio}. More recent frameworks extend this direction with richer temporal structure, broader action spaces, or short-selling support \cite{gu2025mts}. At the optimization level, PPO \cite{schulman2017ppo} and SAC \cite{haarnoja2018sac} remain strong continuous-control baselines and are therefore natural comparators in our setting.

This work also sits alongside classical portfolio allocation and online
portfolio selection. Mean-variance allocation \cite{markowitz1952portfolio},
continuous-time portfolio control \cite{merton1971optimum}, Kelly-style growth
criteria \cite{kelly1956new,busseti2016riskconstrainedkelly}, and universal portfolios \cite{cover1991universal}
all highlight the long-standing trade-off between return, risk, and adaptive
capital allocation. Online multiplicative-update methods further show that
portfolio rules can be updated sequentially as new prices arrive
\cite{helmbold1998line}. PPO-HRAP differs from these approaches because it
learns a policy from simulated trading interaction, but the same trade-off
between upside participation and drawdown control motivates the design.

Among these optimizers, PPO is particularly relevant because it combines policy-gradient learning with a clipped surrogate objective that improves training stability without requiring a fully second-order trust-region method \cite{sutton1999policy,sutton2018reinforcement,schulman2017ppo}. In portfolio allocation problems with continuous actions, this makes PPO a practical backbone for learning bounded position weights while retaining a relatively simple training pipeline \cite{mnih2015human,silver2014deterministic,lillicrap2015continuous}.

Specifically, PPO optimizes the clipped surrogate objective
\begin{equation}
L^{\mathrm{CLIP}}(\theta)=
\mathbb{E}_t\Big[
\min\big(
\kappa_t(\theta)\hat{A}_t,\;
\mathrm{clip}(\kappa_t(\theta),1-\epsilon,1+\epsilon)\hat{A}_t
\big)
\Big],
\end{equation}
where the probability ratio is
\begin{equation}
\kappa_t(\theta)=\frac{\pi_{\theta}(a_t \mid s_t)}{\pi_{\theta_{\mathrm{old}}}(a_t \mid s_t)}.
\end{equation}
The critic estimates the value function $V^\pi(s)$ and the action-value
function
\begin{equation}
Q^\pi(s,a)=\mathbb{E}_\pi\left[\sum_{k=0}^{\infty}\gamma^k r_{t+k}
\mid s_t=s,a_t=a\right],
\end{equation}
which gives the advantage
\begin{equation}
A^\pi(s,a)=Q^\pi(s,a)-V^\pi(s).
\end{equation}
This clipping mechanism constrains overly large policy updates and is one reason PPO remains a practical baseline for financial control tasks with noisy returns.

Another line of work focuses on risk-aware reward engineering. In portfolio problems, performance is commonly summarized with risk-adjusted measures such as the Sharpe ratio \cite{sharpe1994ratio}, Sortino ratio \cite{sortino1994downside}, Calmar ratio, and maximum drawdown. RL trading systems often embed volatility, downside risk, or drawdown penalties into the reward, but reward shaping alone does not necessarily tell the agent which market direction to favor when stress rises \cite{cont2001empirical,engle1982arch,bollerslev1986garch}. PPO-HRAP builds on this observation by retaining drawdown-aware reward shaping while adding an explicit regime-derived action prior.

Drawdown is especially important because it is path dependent. Two policies can
have similar final return but very different investor experience if one reaches
that return through deep interim losses. Prior portfolio literature treats
drawdown and related tail-risk measures as first-class risk objectives
\cite{chekhlov2005drawdown,rockafellar2000cvar}. In an RL environment, however,
path dependence also creates a modeling issue: current drawdown and running
peak must be part of the simulator state used to compute reward. PPO-HRAP uses
these quantities explicitly in the environment and penalizes only drawdown
increases, so the penalty targets worsening losses rather than permanently
punishing a policy for past market stress.

The proposed method also relates to hybrid trading systems that combine learned policies with structured inductive biases. We do not present PPO-HRAP as a purely learned trading rule. It is a hybrid control-plus-RL design in which the final executed action depends on both a deterministic regime prior and the PPO policy output. This framing matters because the reported gains should be interpreted as the joint effect of the prior and the learned correction.

\section{Methodology}
\subsection{Problem formulation}
We formulate daily ETF allocation as a discounted Markov decision process
$\mathcal{M}=(\mathcal{S},\mathcal{A},\mathcal{P},\mathcal{R},\gamma)$. At date
$t$, the agent observes a state $s_t\in\mathcal{S}$, chooses a continuous
portfolio action $a_t\in\mathcal{A}=[-1,1]$, receives reward
$r_t=\mathcal{R}(s_t,a_t,s_{t+1})$, and transitions according to the market and
portfolio simulator $\mathcal{P}$. The discount factor $\gamma$ is inherited
from the PPO training configuration and controls the horizon over which
future trading rewards affect the policy update.

The state is the concatenation of market and portfolio variables:
\begin{equation}
s_t=[x_t^{\mathrm{mkt}},x_t^{\mathrm{port}}]\in\mathbb{R}^{13},
\end{equation}
where $x_t^{\mathrm{mkt}}\in\mathbb{R}^{9}$ and
$x_t^{\mathrm{port}}\in\mathbb{R}^{4}$. The market vector is
\begin{equation}
\begin{split}
x_t^{\mathrm{mkt}}=[&\ell_t,\ \mathrm{sma}_t,\ \mathrm{rsi}_{14,t},\ \mathrm{bb}_t,\
z_t^{\mathrm{vix}},\\
&\mathrm{ret}^{5}_t,\ \mathrm{ret}^{20}_t,\ \mathrm{ma}^{5,20}_t,\
\Delta^{5}\mathrm{vix}_t],
\end{split}
\end{equation}
and the portfolio vector is
\begin{equation}
x_t^{\mathrm{port}}=[c_t,\ w_{t-1},\ \mathrm{upnl}_t,\ dd_t].
\end{equation}
Here $c_t$ is cash ratio, $w_{t-1}$ is the current risky-asset weight,
$\mathrm{upnl}_t=\frac{\mathrm{NAV}_t-\mathrm{NAV}_0}{\mathrm{NAV}_0}$, and
$dd_t$ is current drawdown. The trading environment also tracks date, step,
NAV, running peak, drawdown, current weight, cash ratio, and unrealized PnL
for reward computation and diagnostics.

The inclusion of portfolio variables is important for the Markovian
interpretation of the environment. Market features describe the external price
process, but they do not describe the agent's current exposure. The same price
movement has different consequences when the portfolio is fully invested,
partially in cash, or short. By including current weight, cash ratio,
unrealized PnL, and drawdown, the policy observes the variables that determine
how the next trade will affect NAV and future risk.

\begin{figure}[!t]
\centering
\includegraphics[height=0.51\textheight,keepaspectratio]{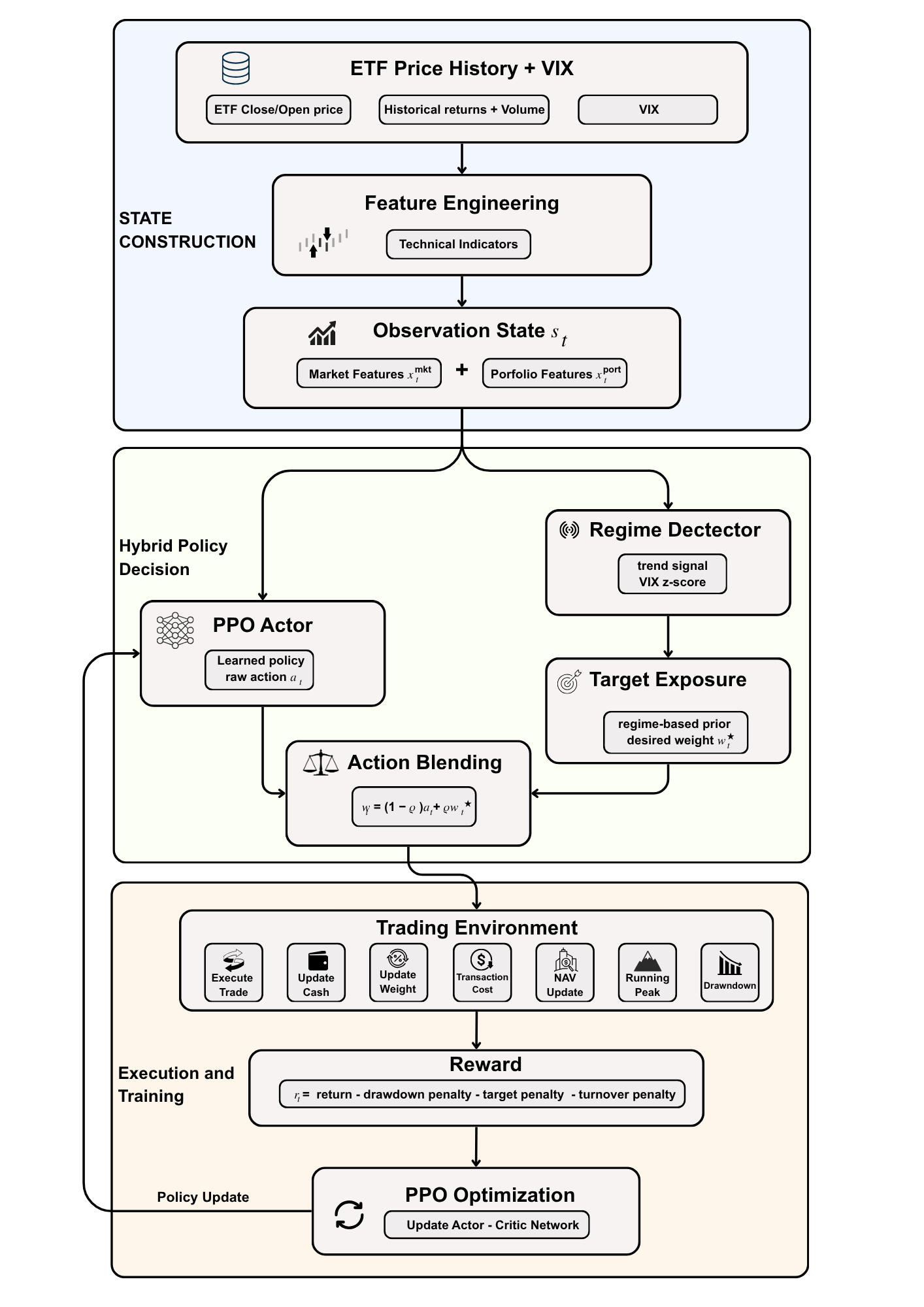}
\caption{PPO-HRAP architecture.}
\label{fig:architecture}
\end{figure}

Figure~\ref{fig:architecture} summarizes the full training-time data flow.
During inference the same feature computation, regime detector, action
blending, and execution steps are used, but the reward is not used to update
the policy. This distinction matters because reward is a learning signal during
training and an evaluation diagnostic during testing, not an input required by
the deployed policy.

\FloatBarrier

\subsection{Feature engineering}
All features are computed from information available before the trading action.
The main formulas are
\begin{align}
\ell_t &= \log\left(\frac{C_t}{C_{t-1}}\right),\\
\mathrm{sma}_t &= \frac{C_t}{\mathrm{SMA}_{50,t}}-1,\\
\mathrm{rsi}_{14,t} &= 100-\frac{100}{1+\mathrm{RS}_{14,t}},\\
\mathrm{bb}_t &= \frac{4\sigma_{20,t}}{\mathrm{SMA}_{20,t}},\\
z_t^{\mathrm{vix}} &= \frac{\mathrm{VIX}_t-\mu_{252,t}}{\sigma_{252,t}},\\
\mathrm{ret}^{k}_t &= \frac{C_t}{C_{t-k}}-1,\quad k\in\{5,20\},\\
\mathrm{ma}^{5,20}_t &= \frac{\mathrm{SMA}_{5,t}}{\mathrm{SMA}_{20,t}}-1,\\
\Delta^5\mathrm{vix}_t &= \frac{\mathrm{VIX}_t}{\mathrm{VIX}_{t-5}}-1.
\end{align}
The RSI definition follows the 14-day relative-strength convention
\cite{wilder1978newconcepts}, while the Bollinger bandwidth uses bands at
$\mathrm{SMA}_{20,t}\pm2\sigma_{20,t}$ \cite{bollinger2002bands}. The
observation includes nine market features and four portfolio features, giving
a 13-dimensional input to the actor.

The feature set intentionally mixes short-horizon and medium-horizon signals.
Daily log return captures immediate price movement; the 5-day and 20-day
returns summarize recent momentum over roughly one trading week and one trading
month; and the moving-average spread gives a compact trend indicator \cite{brock1992simple}. RSI and
Bollinger bandwidth provide complementary information about momentum and
realized dispersion \cite{fischer2018deep}. The VIX z-score and 5-day VIX change are included because
the proposed policy explicitly treats volatility stress as a risk-sizing signal
rather than as a tradable asset.

\subsection{Regime prior and action blending}
The regime detector uses direction and stress signals separately. Direction is
defined by $\mathrm{ret}^{20}_t$ and $\mathrm{ma}^{5,20}_t$: both positive
indicates a bullish regime, while both negative indicates a bearish regime.
Stress is identified from $z_t^{\mathrm{vix}}$: values of at least 0.8 mark a
stressed market and values of at least 1.8 mark a crisis.
These rules map to the target exposure $w_t^\star$ in
Table~\ref{tab:regime-rules}.

\begin{table}[!tb]
\caption{Regime-derived target exposure.}
\label{tab:regime-rules}
\centering
\scriptsize
\begin{tabular}{lcc}
\toprule
Regime & Condition & $w_t^\star$ \\
\midrule
Bull, calm & bull and not stress & 1.0 \\
Bull, stressed & bull and stress & 0.5 \\
Neutral & mixed trend & 0.3 \\
Neutral-negative & weak negative trend & 0.1 \\
Bear, calm & bear without stress & 0.0 \\
Bear, stressed & bear and stress & -0.1 \\
Bear, crisis & bear and crisis & -0.8 \\
\bottomrule
\end{tabular}
\end{table}

The executed portfolio weight is
\begin{equation}
w_t=(1-\rho)a_t+\rho w_t^\star.
\end{equation}
This works as a behavioral prior: the actor can still adapt through $a_t$, but
execution is biased toward a regime-consistent weight. Vanilla PPO must learn
both direction and sizing entirely from delayed reward; PPO-HRAP gives the
policy a stable target when trend and volatility agree.

The blend coefficient $\rho$ controls how strongly this prior affects
execution. When $\rho=0$, the method reduces to executing the raw PPO action.
When $\rho=1$, the learned actor is ignored and the system becomes a
deterministic regime rule. Intermediate values preserve the adaptability of PPO
while preventing the action from drifting too far away from a stress-aware
target. This is the main mechanism by which PPO-HRAP separates the problem of
learning fine-grained allocation corrections from the problem of enforcing a
coarse risk regime.

\subsection{Reward design and optimization}
PPO-HRAP decomposes reward into four interpretable components:
\begin{equation}
r_t=r_t^{\mathrm{return}}-r_t^{\mathrm{drawdown}}-r_t^{\mathrm{target}}-r_t^{\mathrm{turnover}}.
\end{equation}
The return term rewards profitability, the drawdown term discourages worsening
capital loss, the target term encourages regime consistency, and the turnover
term discourages unstable reallocation. In implementation,
\begin{equation}
r_t=\ell_t^p-\lambda_{\mathrm{rrc}}(t)\Delta dd_t
-\beta_{\mathrm{target}}(w_t-w_t^\star)^2-\beta_{\mathrm{turn}}\tau_t,
\end{equation}
where $\ell_t^p=\log\left(\frac{\mathrm{NAV}_{t+1}}{\mathrm{NAV}_t}\right)$ and
$\tau_t=|w_t-w_{t-1}|$.

Each term addresses a different failure mode. The return term prevents the
policy from becoming a purely defensive cash strategy. The drawdown-increase
term penalizes new deterioration relative to the running peak, which is more
targeted than penalizing the absolute drawdown level at every step. The target
penalty discourages the actor from fighting the regime prior when trend and
stress signals agree. The turnover penalty discourages excessive changes in
portfolio weight, which is important because transaction costs are charged at
each reallocation.

The dynamic drawdown coefficient starts from a base penalty and scales it by
market stress:
\begin{equation}
\lambda_{\mathrm{rrc}}(t)=
\max\{0,\lambda_{\mathrm{base}}(1+\alpha\cdot
\mathrm{clip}(z_t^{\mathrm{vix}},-3,3))\}.
\end{equation}
Thus, when VIX z-score rises, the drawdown penalty increases; when stress is
low, the penalty relaxes toward its base level. Drawdown is measured from the
running peak $P_t=\max(P_{t-1},\mathrm{NAV}_t)$:
\begin{equation}
dd_t=\frac{P_t-\mathrm{NAV}_t}{P_t},\qquad
\Delta dd_t=\max(dd_t-dd_{t-1},0).
\end{equation}
The VIX-conditioned coefficient can be interpreted as a first-order stress
adjustment to the base drawdown penalty. Starting from
$\lambda_{\mathrm{base}}$, the multiplier $(1+\alpha z_t^{\mathrm{vix}})$
increases risk aversion when volatility is above its trailing mean and reduces
it when volatility is below normal. Clipping the z-score avoids extreme
coefficients during rare spikes, while the outer maximum prevents the penalty
from becoming negative.

Training uses rewards for policy updates. During inference, the environment
still observes state and executes trades, but reward is only logged for
evaluation; no policy update is performed. Table~\ref{tab:training-loop}
summarizes the training procedure.

\begin{table}[!tb]
\caption{PPO-HRAP training loop (pseudocode).}
\label{tab:training-loop}
\centering
\scriptsize
\begin{tabular}{p{0.92\linewidth}}
\toprule
\textbf{Input:} historical market data, VIX, feature set, $\rho$.\\
Initialize PPO actor-critic parameters.\\
\textbf{for} each training timestep $t$ \textbf{do}\\
\quad Observe $s_t=[x_t^{\mathrm{mkt}},x_t^{\mathrm{port}}]$.\\
\quad Compute regime from trend and VIX z-score.\\
\quad Compute target exposure $w_t^\star$.\\
\quad Sample raw PPO action $a_t$.\\
\quad Blend $w_t=(1-\rho)a_t+\rho w_t^\star$.\\
\quad Execute trade, update NAV, running peak, drawdown.\\
\quad Compute reward $r_t$ and store transition.\\
\quad \textbf{if} the rollout buffer is full \textbf{then} update PPO actor-critic.\\
\textbf{end for}\\
\textbf{Return:} trained policy for inference-time trading.\\
\bottomrule
\end{tabular}
\end{table}

\section{Results}
\subsection{Experimental protocol and baselines}
We use chronological splits: 2010--2017 for training, 2018--2019 for
validation, and 2020--2022 for testing. The SPY split contains 2013, 503, and
756 rows respectively. All methods use a 0.1\% proportional transaction cost.
The main SPY result uses seed 42, the reliability study uses
$\{7,11,19,23,42\}$, and cross-asset transfer is evaluated on QQQ and DIA.

\begin{table}[!tb]
\caption{PPO-HRAP reproducibility settings.}
\label{tab:reproducibility-settings}
\centering
\scriptsize
\begin{tabular}{lr}
\toprule
Parameter & Value \\
\midrule
Total timesteps & 30,000 \\
Main seed & 42 \\
Learning rate & $1\times10^{-4}$ \\
Rollout steps & 1024 \\
Batch size & 128 \\
PPO epochs/update & 10 \\
Entropy coefficient & 0.0 \\
Action-prior weight $\rho$ & 0.6 \\
$\lambda_{\mathrm{base}}$, $\alpha$ & 0.05, 0.5 \\
$\beta_{\mathrm{target}}$, $\beta_{\mathrm{turn}}$ & 0.02, 0.0015 \\
Transaction cost & 0.001 \\
\bottomrule
\end{tabular}
\end{table}

Table~\ref{tab:baseline-formulations} summarizes the benchmark formulations.
Buy and Hold fixes $w_t=1$. Risk parity sets
$w_t=\mathrm{clip}\left(\frac{\sigma^\star}{\hat{\sigma}_{60,t}},0,1\right)$ at a 21-day
rebalance frequency \cite{maillard2008riskparity}. CPPI uses
$w_t=\mathrm{clip}\left(\frac{m(\mathrm{NAV}_t-F)}{\mathrm{NAV}_t},0,1\right)$ with floor
$F=0.8\mathrm{NAV}_0$ and multiplier $m=3$ \cite{black1992cppi}. RL baselines share the same
environment but differ by algorithm or reward.

\begin{table}[!tb]
\caption{Benchmark formulations.}
\label{tab:baseline-formulations}
\centering
\scriptsize
\begin{tabular}{lp{0.62\linewidth}}
\toprule
Method & Formulation \\
\midrule
Buy and Hold & $w_t=1$ after the initial allocation. \\
Risk parity & $w_t=\mathrm{clip}\left(\frac{\sigma^\star}{\hat{\sigma}_{60,t}},0,1\right)$. \\
CPPI & $w_t=\mathrm{clip}\left(\frac{m(\mathrm{NAV}_t-F)}{\mathrm{NAV}_t},0,1\right)$. \\
PPO/SAC profit & $r_t=\ell_t^p$. \\
PPO variance & $r_t=\ell_t^p-\lambda_{\mathrm{var}}\widehat{\mathrm{Var}}_t(\ell^p)$. \\
PPO static MDD & $r_t=\ell_t^p-\lambda_{\mathrm{mdd}}dd_t$. \\
PPO-HRAP & Return, dynamic drawdown, target-consistency, and turnover terms with action blending. \\
\bottomrule
\end{tabular}
\end{table}

\subsection{Main SPY performance}
\begin{figure}[!tb]
\centering
\includegraphics[width=0.96\linewidth]{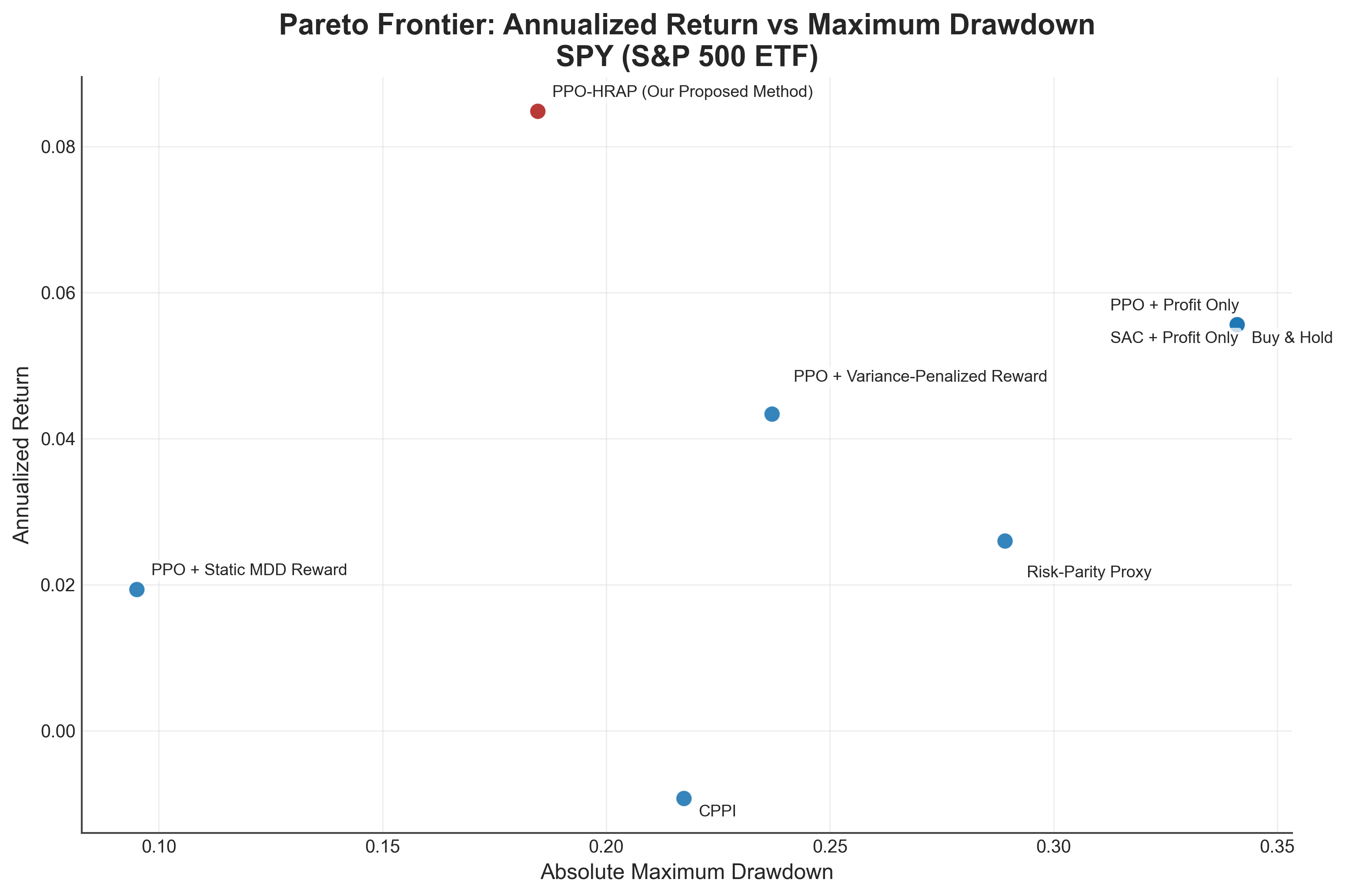}
\caption{SPY return--drawdown trade-off.}
\label{fig:pareto-return-mdd}
\end{figure}

We report total return, annualized return, Sharpe ratio \cite{sharpe1994ratio},
Sortino ratio \cite{sortino1994downside}, Calmar ratio, and maximum drawdown.
Table~\ref{tab:main-results} removes turnover to keep the main comparison
focused on return and risk. PPO-HRAP achieves the strongest total return,
annualized return, Sharpe, Sortino, and Calmar on SPY. Relative to Buy and
Hold, Sharpe improves from 0.3420 to 0.6447 and maximum drawdown decreases
from 34.10\% to 18.47\%. The smallest drawdown is obtained by PPO Static MDD
(9.51\%), but at the cost of a much lower total return (5.91\%).
Figure~\ref{fig:pareto-return-mdd} visualizes this risk-return trade-off
directly by comparing total return against maximum drawdown.

\begin{table*}[!t]
\caption{Held-out test performance on SPY, 2020--2022. Best values are bold.}
\label{tab:main-results}
\centering
\scriptsize
\begin{tabular}{lrrrrrr}
\toprule
Method & Total Return & Ann. Return & Sharpe & Sortino & Calmar & MDD \\
\midrule
Buy and Hold & 0.1760 & 0.0556 & 0.3420 & 0.4251 & 0.1630 & -0.3410 \\
Risk-Parity & 0.0799 & 0.0260 & 0.2281 & 0.2848 & 0.0898 & -0.2892 \\
CPPI & -0.0275 & -0.0093 & -0.0021 & -0.0026 & -0.0427 & -0.2174 \\
PPO Profit & 0.1760 & 0.0556 & 0.3420 & 0.4251 & 0.1630 & -0.3410 \\
SAC Profit & 0.1760 & 0.0556 & 0.3420 & 0.4251 & 0.1630 & -0.3410 \\
PPO Variance & 0.1356 & 0.0434 & 0.3384 & 0.4208 & 0.1829 & -0.2371 \\
PPO Static MDD & 0.0591 & 0.0193 & 0.3341 & 0.4153 & 0.2034 & \textbf{-0.0951} \\
PPO-HRAP & \textbf{0.2762} & \textbf{0.0848} & \textbf{0.6447} & \textbf{0.8588} & \textbf{0.4592} & -0.1847 \\
\bottomrule
\end{tabular}
\end{table*}

\begin{figure*}[!t]
\centering
\includegraphics[width=0.84\textwidth]{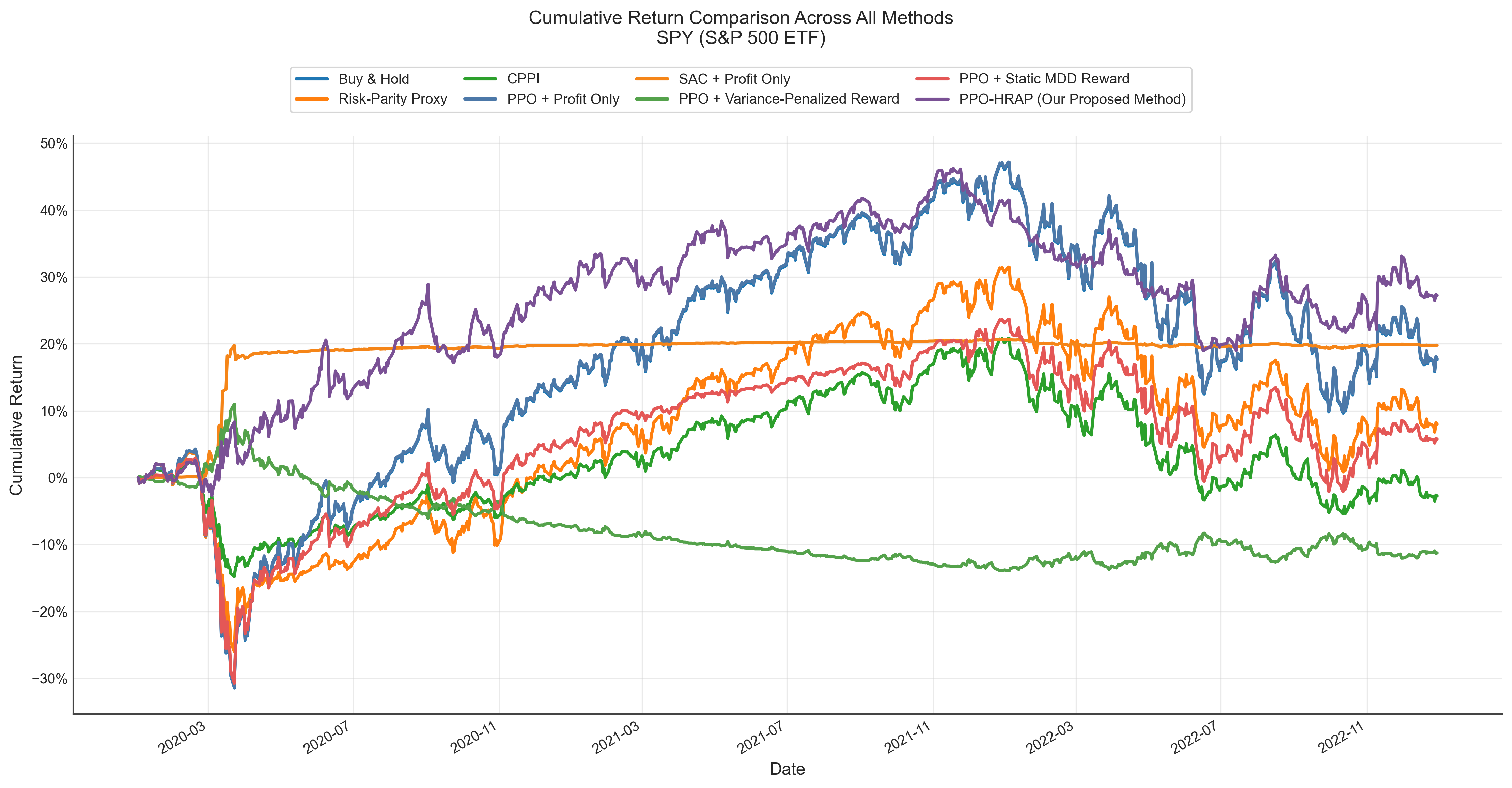}
\caption{SPY cumulative returns across methods.}
\label{fig:cumulative-spy}
\end{figure*}

\subsection{SPY multi-seed reliability}
Single-run trading results can be sensitive to initialization, especially for
policy-gradient methods trained on noisy financial rewards. To check whether
the proposed behavior is tied to one favorable random seed, we rerun the RL
methods on seeds $\{7,11,19,23,42\}$. Table~\ref{tab:multiseed-results}
reports the mean and standard deviation of the main risk-return metrics.

\begin{table*}[!t]
\caption{SPY multi-seed test summary across RL methods.}
\label{tab:multiseed-results}
\centering
\scriptsize
\begin{tabular}{lrrrrrrrr}
\toprule
Method & Ret. Mean & Ret. Std & Sharpe Mean & Sharpe Std & Calmar Mean & Calmar Std & MDD Mean & MDD Std \\
\midrule
PPO-HRAP & \textbf{0.2725} & 0.0109 & 0.6219 & 0.0565 & 0.4508 & 0.0499 & -0.1876 & 0.0210 \\
PPO Static MDD & 0.0575 & 0.2008 & 0.1320 & 0.3882 & 0.0550 & 0.1829 & -0.3554 & 0.0321 \\
PPO Profit & 0.1781 & \textbf{0.0059} & 0.3444 & \textbf{0.0067} & 0.1648 & \textbf{0.0053} & -0.3413 & \textbf{0.0005} \\
PPO Variance & -0.1126 & 0.1062 & -0.3720 & 0.2117 & -0.1512 & 0.0851 & -0.2264 & 0.1560 \\
SAC Profit & 0.1978 & 0.0897 & \textbf{0.8644} & 0.2417 & \textbf{1.4733} & 0.6717 & \textbf{-0.0469} & 0.0199 \\
\bottomrule
\end{tabular}
\end{table*}

The multi-seed results show two useful nuances. First, PPO-HRAP has the best
mean total return among the RL methods and a small return standard deviation,
indicating that the performance improvement is not isolated to seed 42.
Second, SAC profit has stronger mean Sharpe, Calmar, and drawdown in this
multi-seed summary, but it also produces lower mean total return than PPO-HRAP.
We therefore interpret the proposed method as a return-leading hybrid policy
with improved drawdown control relative to passive long exposure, not as a
universal winner on every risk metric.

\subsection{Cross-asset evaluation}
\begin{figure*}[!t]
\centering
\includegraphics[width=0.92\textwidth]{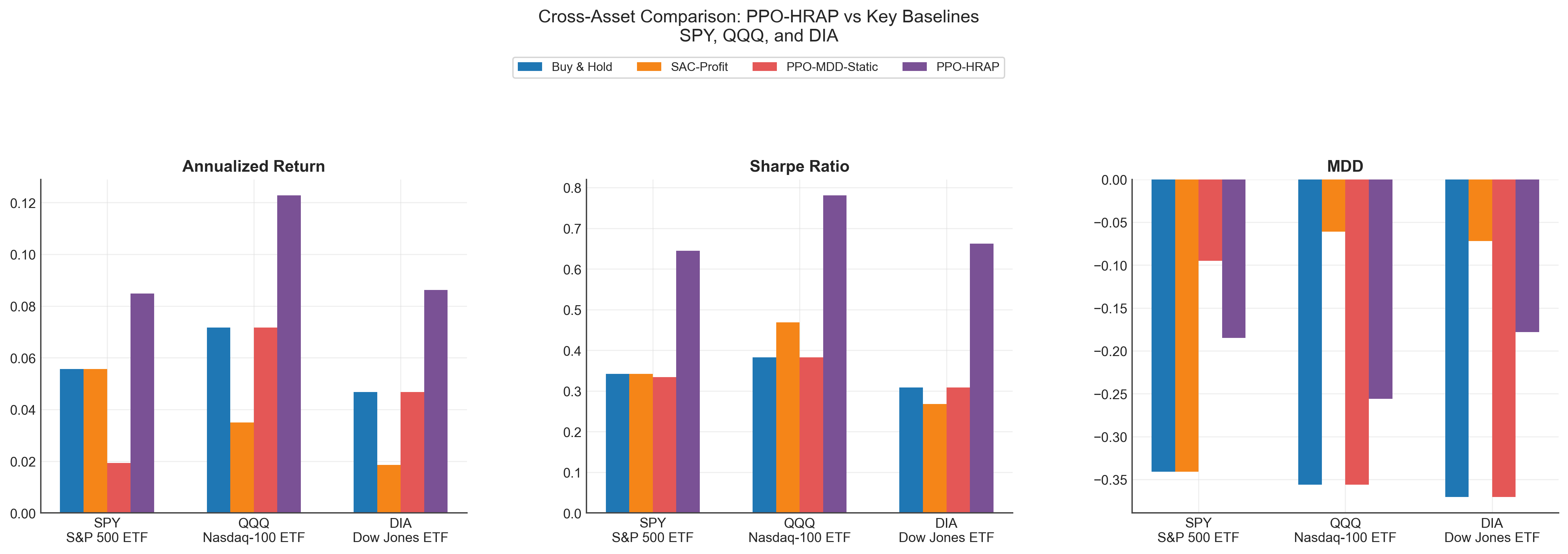}
\caption{Cross-asset performance on SPY, QQQ, and DIA.}
\label{fig:cross-bars}
\end{figure*}

\begin{figure}[!tb]
\centering
\includegraphics[width=0.96\linewidth]{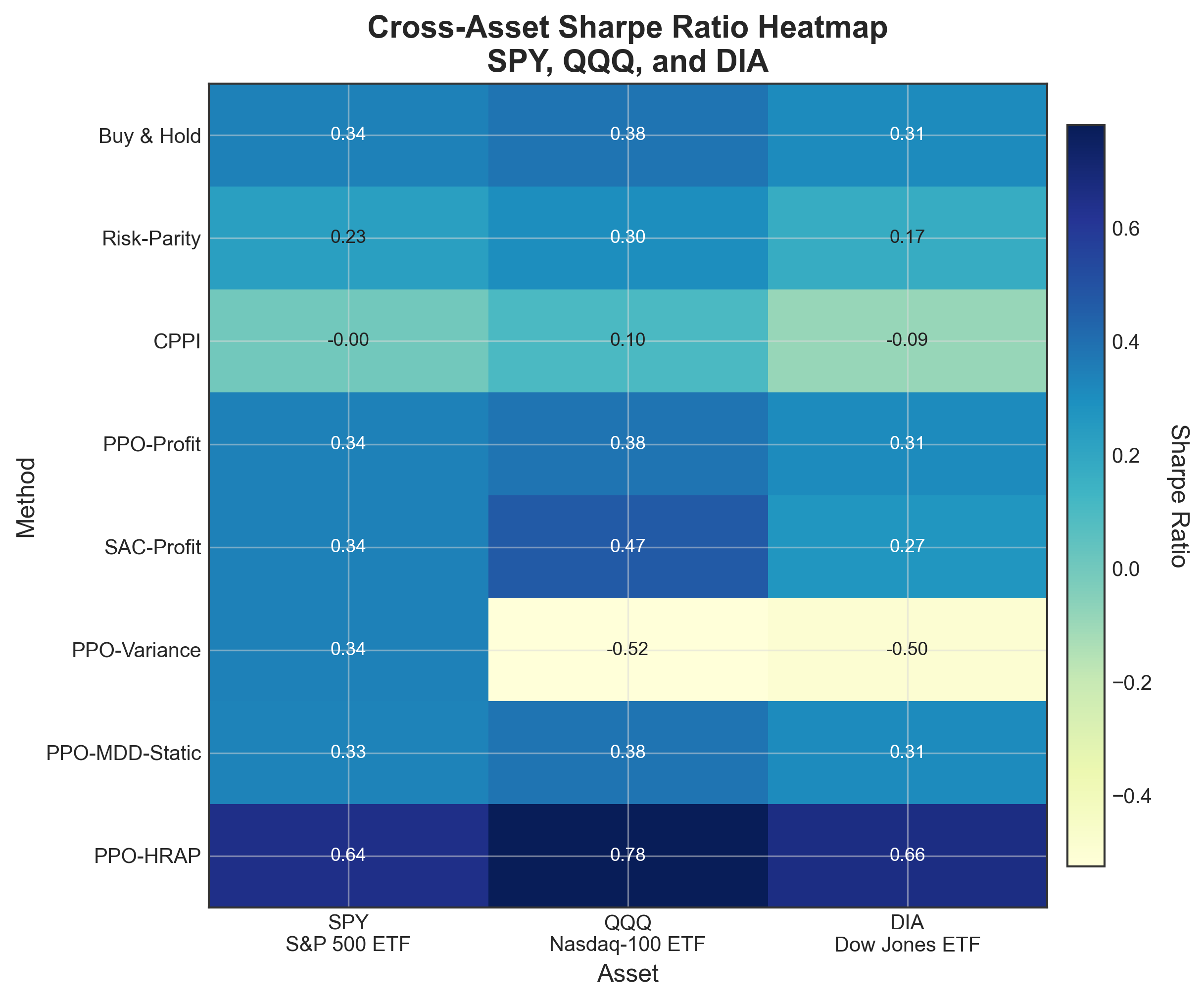}
\caption{Cross-asset Sharpe-ratio heatmap.}
\label{fig:cross-heatmap}
\end{figure}

The cross-asset runs test whether the regime-guided prior transfers beyond
SPY. PPO-HRAP ranks first on total return and Sharpe ratio for SPY, QQQ, and
DIA. It also ranks first on Calmar ratio for SPY and DIA; on QQQ, SAC profit
has the best Calmar because it produces a smaller drawdown. Figure~\ref{fig:cross-bars}
compares PPO-HRAP with key baselines, and Figure~\ref{fig:cross-heatmap}
shows the Sharpe-ratio matrix across all methods and assets.

Table~\ref{tab:qqq-dia-test-results} reports the full single-run QQQ and DIA
test metrics across the compared methods.

\begin{table*}[!t]
\caption{QQQ and DIA single-run test performance across methods.}
\label{tab:qqq-dia-test-results}
\centering
\scriptsize
\begin{tabular}{llrrrrrrr}
\toprule
Asset & Method & Total Ret. & Ann. Ret. & Sharpe & Sortino & Calmar & MDD & Turnover \\
\midrule
QQQ & Buy and Hold & 0.2306 & 0.0717 & 0.3826 & 0.5082 & 0.2014 & -0.3562 & 1.0000 \\
QQQ & Risk Parity & 0.1317 & 0.0422 & 0.2985 & 0.3954 & 0.1512 & -0.2788 & 3.8472 \\
QQQ & CPPI & -0.0072 & -0.0024 & 0.1010 & 0.1361 & -0.0067 & -0.3605 & 9.7168 \\
QQQ & PPO Static MDD & 0.2306 & 0.0717 & 0.3826 & 0.5082 & 0.2014 & -0.3562 & 1.0000 \\
QQQ & PPO Profit & 0.2306 & 0.0717 & 0.3826 & 0.5082 & 0.2014 & -0.3562 & 1.0000 \\
QQQ & PPO Variance & -0.1770 & -0.0630 & -0.5244 & -0.8868 & -0.1951 & -0.3227 & 35.3054 \\
QQQ & SAC Profit & 0.1084 & 0.0350 & 0.4684 & 0.6231 & 0.5739 & -0.0609 & 10.0306 \\
QQQ & PPO-HRAP & 0.4149 & 0.1228 & 0.7811 & 1.0697 & 0.4799 & -0.2559 & 39.4215 \\
\midrule
DIA & Buy and Hold & 0.1468 & 0.0468 & 0.3088 & 0.3675 & 0.1263 & -0.3706 & 1.0000 \\
DIA & Risk Parity & 0.0441 & 0.0145 & 0.1727 & 0.2107 & 0.0482 & -0.3013 & 2.7133 \\
DIA & CPPI & -0.0407 & -0.0138 & -0.0900 & -0.1085 & -0.0813 & -0.1696 & 7.3029 \\
DIA & PPO Static MDD & 0.1468 & 0.0468 & 0.3088 & 0.3675 & 0.1263 & -0.3706 & 1.0000 \\
DIA & PPO Profit & 0.1468 & 0.0468 & 0.3088 & 0.3675 & 0.1263 & -0.3706 & 1.0000 \\
DIA & PPO Variance & -0.1266 & -0.0442 & -0.4967 & -0.5515 & -0.2245 & -0.1967 & 52.6533 \\
DIA & SAC Profit & 0.0568 & 0.0186 & 0.2677 & 0.2981 & 0.2586 & -0.0720 & 7.2781 \\
DIA & PPO-HRAP & 0.2811 & 0.0862 & 0.6623 & 0.8622 & 0.4838 & -0.1782 & 55.5914 \\
\bottomrule
\end{tabular}
\end{table*}

\subsection{Action-prior sensitivity and behavior}
The validation search over $\rho$ selects $\rho=0.6$, with validation Calmar
0.2299 and validation total return 0.0800. Performance is not monotonic in
$\rho$: a weak prior does not sufficiently regularize the actor, while an
overly strong prior can suppress learned correction. On
SPY test data, PPO-HRAP has average high-VIX exposure 0.4337 and average
low-VIX exposure 0.8839, confirming that the policy lowers exposure under
stress. It still spends 54.10\% of the test period at full-long exposure, so
the prior mitigates but does not eliminate long bias.

\begin{table}[!tb]
\caption{Validation sweep for action-prior weight $\rho$ on SPY.}
\label{tab:rho-sensitivity}
\centering
\scriptsize
\begin{tabular}{rrrrrl}
\toprule
$\rho$ & Return & Sharpe & Calmar & MDD & Filter \\
\midrule
0.3 & 0.0686 & 0.3770 & 0.1792 & -0.1889 & no \\
0.4 & 0.0729 & 0.4146 & 0.1941 & -0.1854 & no \\
0.5 & 0.0771 & 0.4537 & 0.2126 & -0.1787 & yes \\
0.6 & \textbf{0.0800} & \textbf{0.4863} & \textbf{0.2299} & -0.1713 & yes \\
0.7 & 0.0767 & 0.4835 & 0.2291 & \textbf{-0.1650} & yes \\
\bottomrule
\end{tabular}
\end{table}

\begin{figure}[!tb]
\centering
\includegraphics[width=0.98\linewidth]{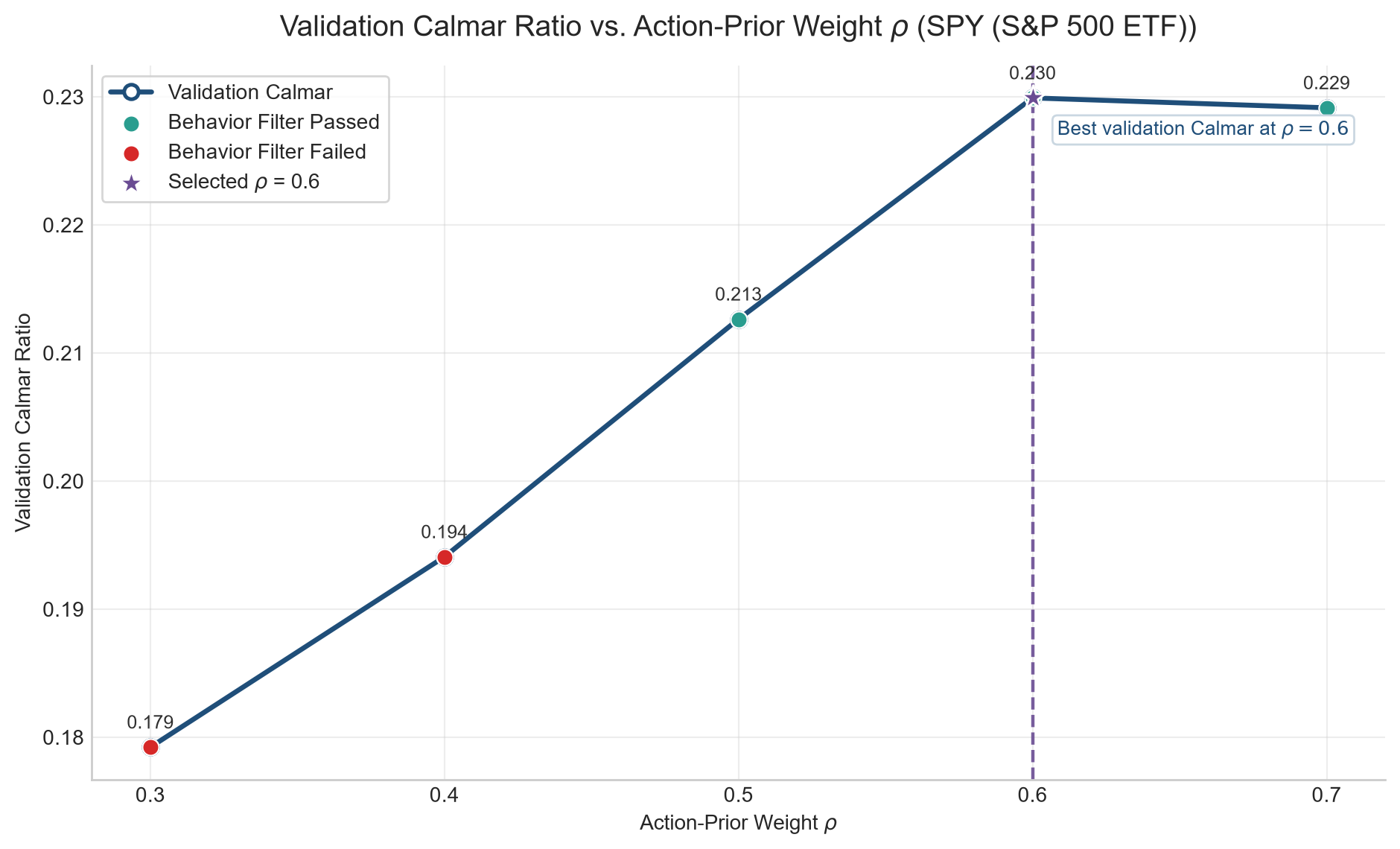}
\caption{Validation Calmar over action-prior weight $\rho$.}
\label{fig:rho-sensitivity}
\end{figure}

\section{Discussion}
The main benefit of PPO-HRAP appears to come from decoupling directional guidance from risk sizing. Reward-only baselines either behave nearly identically to Buy and Hold or become too conservative. By blending the learned action with a regime-derived target, PPO-HRAP injects a strong inductive bias that remains consistent with market stress and trend direction while still allowing PPO to fine-tune execution.

The behavioral diagnostics also help explain why PPO-HRAP improves the aggregate metrics without eliminating long exposure entirely. Under high-VIX conditions, average exposure falls from 0.8839 in calmer periods to 0.4337, which indicates that the hybrid policy does react to stress rather than staying mechanically fully invested. At the same time, the method still spends 54.10\% of the test period at full long exposure, so the current regime prior should be viewed as a partial correction to vanilla PPO behavior rather than a complete solution to long bias.

This practical benefit comes with clear trade-offs. First, turnover is high: PPO-HRAP reaches 51.22 on the SPY test split and remains around 39--56 across SPY, QQQ, and DIA, far above passive or reward-only baselines. Second, although the five-seed SPY study substantially strengthens the reliability story, the cross-asset evidence on QQQ and DIA is still single-run and therefore should be interpreted more cautiously. Third, the training budget remains modest at 30,000 timesteps, so the reported results are best viewed as a strong benchmark snapshot rather than a fully saturated optimization study. The paper therefore supports PPO-HRAP as a promising engineering design, but not yet as a universally validated trading policy.

\section{Conclusion}
This paper presented PPO-HRAP, a PPO-based hybrid regime-aware policy for risk-controlled trading. The method combines dynamic drawdown control, regime-guided target exposure, and action blending to address the limitations of reward-only RL trading on upward-drifting equity ETFs. In the main 2020--2022 SPY run, PPO-HRAP improves total return and risk-adjusted metrics over both financial and RL baselines while substantially reducing maximum drawdown relative to Buy and Hold. The multi-seed SPY results show that these gains are not confined to a single random seed, and the single-run transfers to QQQ and DIA indicate that the hybrid regime prior can remain useful beyond the original market.

The current findings support the practical value of hybrid control-plus-learning policies in financial RL. The most important next steps are broader multi-seed cross-asset evaluation, longer training budgets, deeper turnover analysis, richer regime definitions, and broader testing across additional assets and market conditions.

\section*{Code Availability}
The implementation is available at
\url{https://github.com/chikien07012006/PPO_Regime-Aware-Trading}.

\bibliographystyle{ACM-Reference-Format}
\bibliography{refs}

\end{document}